\documentclass[11pt,a4paper]{article}
\usepackage[hyperref]{emnlp2018}
\usepackage{times}
\usepackage{latexsym}
\usepackage{amsmath}
\usepackage{url}
\usepackage{graphicx}
\usepackage{subcaption}
\usepackage{amssymb}
\usepackage{multirow}
\usepackage{arydshln}
\def\code#1{\texttt{#1}}

\aclfinalcopy

 \title{Beyond Weight Tying: Learning Joint Input-Output Embeddings \\ for  Neural Machine Translation}
 \author{Nikolaos Pappas$^\dagger$ \ \ \ \  Lesly Miculicich Werlen$^\dagger$$^\diamondsuit$ \ \ \ James Henderson$^\dagger$\\
 \textsuperscript{$\dagger$}{Idiap Research Institute, Martigny, Switzerland} \\
\textsuperscript{$\diamondsuit$}{\'{E}cole Polytechnique F\'{e}d\'{e}rale de Lausanne (EPFL), Switzerland}\\
	{\tt \{npappas,lmiculicich,jhenderson\}@idiap.ch}
}

\date{}

\begin{document}
\maketitle

\begin{abstract}
	Tying the weights of the target word embeddings with the target word classifiers of neural machine translation models leads to faster training and often to better translation quality. Given the success of this parameter sharing, we investigate other forms of sharing in between no sharing and hard equality of parameters.  In particular, we propose a \textit{structure-aware} output layer which
  captures the semantic structure of the output space of words within a joint input-output embedding. The model is a  generalized form of \textit{weight tying}
  which shares parameters but allows learning a more flexible relationship with input word embeddings and allows the effective capacity of the output layer to be controlled.
  In addition, the model shares weights across output classifiers and translation contexts which allows it to better leverage prior knowledge about them.
  Our evaluation on English-to-Finnish and English-to-German datasets shows the effectiveness of the method against strong encoder-decoder baselines trained with or without \textit{weight tying}.
\end{abstract}

\section{Introduction}
Neural machine translation (NMT) predicts the target sentence one word at a time, and thus models the task as a  sequence classification problem where the classes correspond to words. Typically, words are treated as categorical variables which lack description and semantics. This makes training speed and parametrization dependent on the size of the target vocabulary \cite{mikolov2013distributed}. Previous studies  overcome this problem by truncating the vocabulary to limit its size and mapping out-of-vocabulary words to a single ``unknown'' token. Other approaches attempt to use a limited number of frequent words plus \emph{sub-word units} \cite{sennrich-haddow-birch:2016:P16-12}, the combination of which can cover the full vocabulary, or to perform character-level modeling \cite{chung-cho-bengio:2016:P16-1, TACL1051, costajussa-fonollosa:2016:P16-2, ling2015character}; with the former being the most effective between the two. The idea behind these alternatives is to overcome the vocabulary size issue by modeling the morphology of rare words. One  limitation, however, is that semantic information of words or sub-word units learned by the input embedding are not considered when learning to predict output words. Hence, they rely on a large amount of examples per class to learn proper word or sub-word unit output classifiers.

\begin{table*}
	\centering
	{\def\arraystretch{0.98}\tabcolsep=10pt
\begin{tabular}{ c c  c c cc }
  \hline
   & \multicolumn{2}{c}{NMT} & NMT-\code{tied} & \multicolumn{2}{c}{NMT-\code{joint}} \\
	\textbf{Query} & \textbf{Input} & \textbf{Output} & \textbf{Input/Output} & \textbf{Input} & \textbf{Output} \\

  \hline
  visited & attacked  & \textcolor{red}{visiting}  & \textcolor{red}{visits}    &  visiting  & attended \\
		(Verb past tense)			& conquered  & attended & attended &
attended  & witnessed\\
   			  & contacted & \textcolor{red}{visit}  & \textcolor{red}{visiting}  &
visits  & discussed \\
 					& occupied & \textcolor{red}{visits} & frequented & visit & recognized \\
					& consulted  & discovered & \textcolor{red}{visit} &
frequented  & demonstrated \\

\hdashline
generous & modest  & spacious  &  \textcolor{red}{generosity}     &  spacious &  friendly\\
	(Adjective)			& extensive  & \textcolor{red}{generosity} & spacious &
generosity  & flexible\\
				& substantial & \textcolor{red}{generously}  & \textcolor{red}{generously}   &
flexible  & brilliant \\
				& ambitious & massive & lavish & generously & fantastic \\
				& sumptuous  & huge & massive &
massive  & massive \\\hdashline

friend & wife  & \textcolor{red}{friends} & colleague & colleague  & colleague  \\
	(Noun)	 & husband  & colleague  & \textcolor{red}{friends}  & friends  & fellow  \\
		 & colleague & \textcolor{red}{Fri@@}  & neighbour  & neighbour  & supporter\\
		 & friends & fellow & girlfriend  & girlfriend  & partner \\
		 & painter & \textcolor{red}{friendship} & companion  & husband & manager \\ \hline

  \hline
\end{tabular}}
\caption{Top-5 most similar input and output representations to two query words based on cosine similarity for an NMT trained without (NMT) or with \textit{weight tying} (NMT-\code{tied}) and our \textit{structure-aware} output layer (NMT-\code{joint}) on De-En ($|\mathcal{V}| \approx 32K$). Our model learns representations useful for encoding and generation which are more consistent to the dominant semantic and syntactic relations of the query such as verbs in past tense, adjectives and nouns (inconsistent words are marked in \textcolor{red}{red}).}
\label{example}
\end{table*}
One way to consider information learned by input embeddings, albeit restrictively, is with \textit{weight tying} i.e.~sharing the parameters of the input embeddings with those of the output classifiers \cite{press-wolf:2017:EACLshort, inan2016tying} 
which is effective for language modeling and machine translation  \cite{sennrich-EtAl:2017:EACLDemo, klein-EtAl:2017:ACL-2017-System-Demonstrations}.
Despite its usefulness,  we find that \textit{weight tying} has three limitations:
(a) It biases all the words with similar input embeddings to have a similar chance to be generated, which may not always be the case (see Table \ref{example} for examples).  
Ideally, it would be better to learn distinct relationships  useful for encoding and decoding 
without forcing any general bias. 
(b) The relationship between outputs is only implicitly captured by \textit{weight tying} because there is no parameter sharing across output classifiers.
(c) It requires that the size of the translation context vector and the input embeddings are the same, which in practice makes it difficult to control the output layer capacity.

In this study, we propose a \textit{structure-aware} output layer which overcomes the limitations of previous output layers of NMT models. To achieve this, we treat words and subwords as units with textual descriptions and semantics. 
The model consists of a joint input-output embedding which learns what to share between input embeddings and output classifiers, but also shares parameters across output classifiers and translation contexts to better capture the similarity structure of the output space and leverage prior knowledge about this similarity.  This flexible sharing allows it to distinguish between features of words which are useful for encoding, generating, or both. Figure \ref{example} shows examples of the proposed model's input and output representations, compared to those of a softmax linear unit with or without \textit{weight tying}.

  This proposal is inspired by joint input-output models for zero-shot text classification \cite{yazdani2015model,nam2016all}, but innovates in three important directions, namely  in learning complex non-linear relationships, controlling the effective capacity of the output layer and handling structured prediction problems. 

Our contributions are summarized as follows:
\begin{itemize}
	\item We identify key theoretical and practical limitations of existing output layer parametrizations such as softmax linear units with or without \textit{weight tying} and relate the latter to joint input-output models.
	\item We propose a novel \textit{structure-aware} output layer which has flexible  parametrization for neural MT and demonstrate that its mathematical form is a generalization of existing output layer parametrizations.
	\item We provide empirical evidence of the superiority of the proposed structure-aware output layer on morphologically simple and  complex languages as targets, including under challenging conditions, namely varying vocabulary sizes, architecture depth, and output frequency.
\end{itemize}
\vspace{-2mm}
The evaluation is performed on 4 translation pairs, namely English-German and English-Finnish in both directions using BPE \cite{sennrich-haddow-birch:2016:P16-12} of varying operations to investigate the effect of the vocabulary size to each model.
The main baseline is a strong LSTM encoder-decoder model with 2 layers on each side (4 layers) trained with or without \textit{\textit{weight tying}} on the target side, but we also experiment with deeper models with up to 4 layers on each side (8 layers).
To improve efficiency on large vocabulary sizes we make use of negative sampling as in \cite{mikolov2013distributed} and show that the proposed model is the most robust to such approximate training among the alternatives.

\section{Background: Neural MT}
\label{lanmt}
The translation objective is to maximize the conditional probability of emitting a sentence in a target language \(Y =\{y_1,...,y_n\}\)  given a sentence in a source language \(X =\{x_1,...,x_m\}\), noted  \(p_{\Theta}(Y | X)\), where \(\Theta\) are the model parameters learned from a parallel corpus of length $N$:
\vspace{-2mm}
\begin{equation}
 \max_{\Theta} \frac{1}{N} \sum_{i=1}^N \log(p_{\Theta}(Y^{(i)}|X^{(i)})).
\end{equation}
By applying the chain rule, the output sequence can be generated one word at a time by calculating the following conditional distribution:
\begin{equation}
p(y_t|y_1^{t-1},X ) \approx f_{\Theta}(y_1^{t-1},X).
\end{equation}
where $f_{\Theta}$ returns a column vector with an element for each $y_t$.
Different models have been proposed to approximate the function $f_{\Theta}$ \cite{kalchbrenner13,Sutskever14,bahdanau2014neural,cho-EtAl:2014:EMNLP2014, gehring2017convolutional, vaswani2017attention}. Without loss of generality, we focus here on LSTM-based encoder-decoder model with  attention
\citet{luong-pham-manning:2015:EMNLP}.

\subsection{Output Layer parametrizations}
\subsubsection{Softmax Linear Unit}
The most common output layer (Figure \ref{fig:tied}), consists of a linear unit with a weight matrix $W\in \rm I\!R^{d_h \times |\mathcal{V}|}$ and a bias vector  $b  \in \rm I\!R^{|\mathcal{V}|}$ followed by a softmax activation function,  where $V$ is the vocabulary, noted as NMT. For brevity, we focus our analysis specifically on the nominator of the normalized exponential which characterizes softmax. Given the decoder's hidden representation ${h}_t$ with dimension size $d_h$, the output probability distribution at a given time, $y_t$, conditioned on the input sentence $X$ and the previously predicted outputs $y_1^{t-1}$ can be written as follows:
\begin{align}
 p(y_t|y_1^{t-1},X) & \propto  \text{exp}(W^T{h}_t + b) \nonumber  \\
          & \propto \text{exp}(W^TI{h}_t + b),
          \label{baseline_eq}
\end{align}
where $I$ is the identity function. From the second line of the above equation, we observe that there is no explicit output space structure learned by the model because there is no parameter sharing across outputs;~the parameters for output class $i$,  $W^T_i$, are independent from parameters for any other output class $j$, $W^T_j$.

\subsubsection{Softmax Linear Unit with \textit{Weight Tying}}
\label{weighttying}
The parameters of the output embedding $W$ can be tied with the parameters of the input embedding $E \in \rm I\!R^{|\mathcal{V}| \times d}$ by setting $W = E^T$, noted as NMT-\code{tied}. This can happen only when the input dimension of $W$ is restricted to be the same as that of the input embedding ($d = d_h$). This creates practical limitations because the optimal dimensions of the input embedding and translation context may actually be when $d_h \neq d$. 

With tied embeddings, the parametrization of the conditional output probability distribution from Eq.~\ref{baseline_eq} can be re-written as:
\vspace{-3mm}
\begin{align}
  p(y_t|y_1^{t-1},X) & \propto    \text{exp}((E^T)^T{h}_t + b) \nonumber \\
  & \propto  \text{exp}(E{h}_t + b).
	\label{tied_eq}
\end{align}
As above, this model does not capture any explicit output space structure. However, previous studies have shown that the input embedding learns linear relationships between words similar to distributional methods \cite{mikolov2013distributed}. The hard equality of parameters imposed by $W=E^T$ forces the model to re-use this implicit structure in the output layer and increases the modeling burden of the decoder itself by requiring it to match this structure through $h_t$. Assuming that the latent linear structure which E learns is of the form $E \approx E_{l} \mathcal{W}$ where $E_{l} \in \rm  I\!R^{|\mathcal{V}| \times k}$ and $\mathcal{W} \in \rm I\!R^{k \times d}$ and $d = d_h$, then Eq.~\ref{tied_eq} becomes:
\begin{equation}
\begin{split}
       p(y_t|y_1^{t-1},X) &  \propto exp(E_{l} \mathcal{W} {h}_t + b)\ \square.
\end{split}
\end{equation}

\begin{figure*}[htbp]
  \centering
    \begin{subfigure}[b]{0.45\textwidth}
      \centering
        \includegraphics[width=0.75\textwidth]{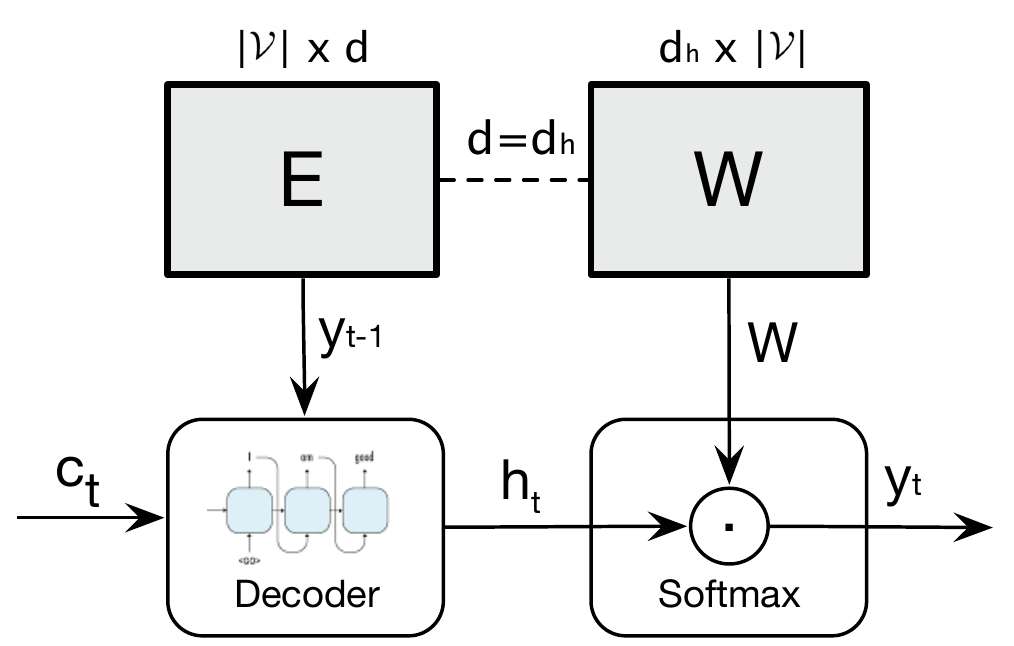}
        \vspace{-0mm}\caption{Typical output layer which is a softmax linear unit without or with \textit{\textit{weight tying}} ($W=E^T$).}
        \label{fig:tied}
    \end{subfigure}
		\hspace{3mm}
    \begin{subfigure}[b]{0.47\textwidth}
      \centering
        \includegraphics[width=0.9\textwidth]{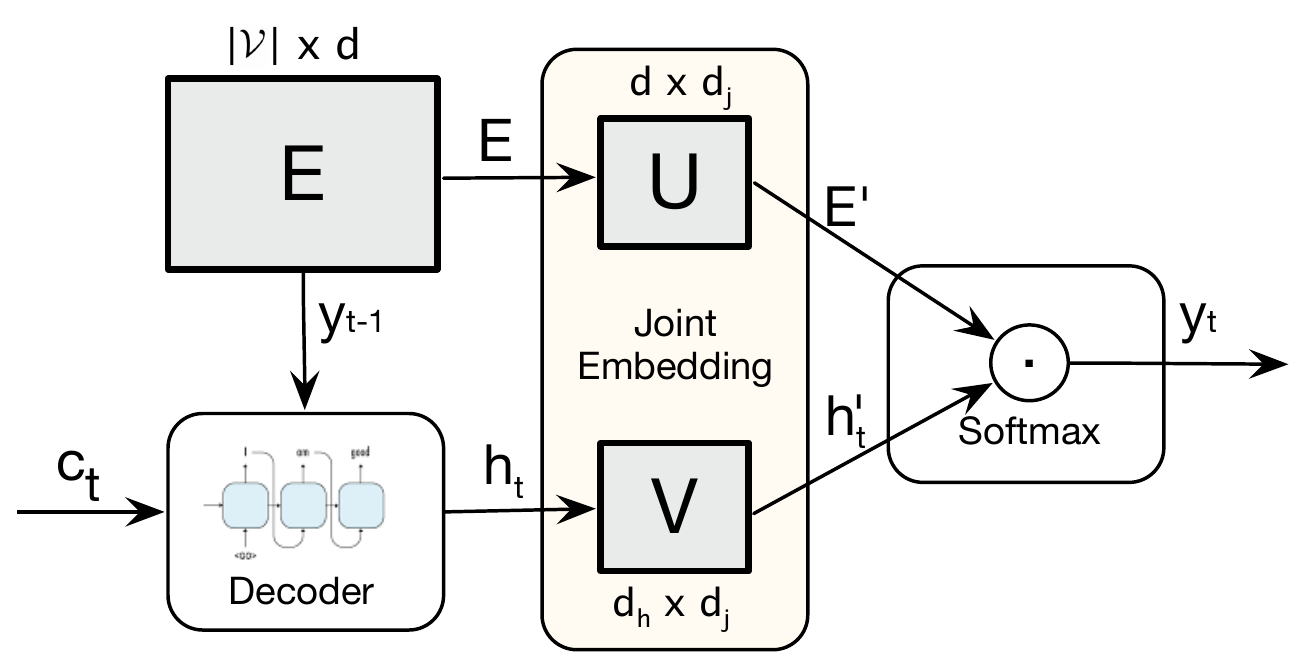}
        \vspace{-0mm}\caption[1]{The \textit{structure-aware} output layer is  a joint embedding between translation contexts and word classifiers. }
        \label{fig:proposed}
    \end{subfigure}
    \caption[1]{Schematic of existing output layers and the proposed output layer for the decoder of the NMT model with source context vector $c_t$, previous word $y_{t-1} \in \rm I\!R^d$, and decoder hidden states, $h_t \in \rm I\!R^{d_h}$. }
\end{figure*}
The above form, excluding bias $b$, shows that \textit{weight tying} learns a similar linear structure, albeit implicitly, to joint input-output embedding models with a bilinear form for zero-shot classification \cite{yazdani2015model,nam2016all}.\footnote{The capturing of implicit structure could also apply for the output embedding $W$ in Eq.~\ref{baseline_eq}, however that model would not match the bilinear input-output model form because it is based on the input embedding $E$.}  This may explain why \textit{weight tying} is more sample efficient than the baseline softmax linear unit, but also motivates the learning of explicit structure through joint input-output models.

\subsection{Challenges}
\label{challenges}
We identify two key challenges of the existing parametrizations of the output layer: (a) their difficulty in learning complex structure of the output space due to their bilinear form and (b) their rigidness in controlling the output layer capacity due to their strict equality of the dimensionality of the translation context and the input embedding.

\subsubsection{Learning Complex Structure}
The existing joint input-output embedding models \cite{yazdani2015model,nam2016all} have the following bilinear form:
\begin{align}
  E \hspace{-0.5ex}\underbrace{\mathcal{W}}_{\text Structure}\hspace{-0.5ex} h_{t}\,
  \label{eq:bilinear}
\end{align}
where $\mathcal{W} \in \rm I\!R^{d \times d_h}$. We can observe that the above formula can only capture linear relationships between encoded text ($h_t$) and input embedding ($E$) through $\mathcal{W}$. We argue that for structured prediction, the relationships between different outputs are more complex due to complex interactions of the semantic and syntactic relations across outputs but also between outputs and different contexts. A more appropriate form for this purpose would include a non-linear transformation $\sigma(\cdot)$, for instance with either:
\begin{align}
  \text{(a)} \underbrace{\sigma(E \mathcal{W})}_{\text{Output structure}}\hspace{-1ex}  h_{t}\ \
  \text{~~or~~}\ \ \text{(b)~~} E \hspace{-1ex}\underbrace{\sigma(\mathcal{W} h_{t})}_{\text{Context structure}}.
	\label{nonlin}
\end{align}
\subsubsection{Controlling Effective Capacity}
Given the above definitions we now turn our focus to a more practical challenge, which is the capacity of the output layer. Let $\Theta_{base}$, $\Theta_{tied}$, $\Theta_{bilinear}$ be the parameters associated with a softmax linear unit without and with \textit{weight tying} and with a joint bilinear input-output embedding, respectively. The capacity of the output layer in terms of effective number of parameters can be expressed as:
\begin{align}
		\mathcal{C}_{base} \approx |\Theta_{base}| = |\mathcal{V}| \times d_h + |\mathcal{V}|\\
		\mathcal{C}_{tied} \approx |\Theta_{tied}| \leq |\mathcal{V}| \times d_h + |\mathcal{V}|\\
		\mathcal{C}_{bilinear} \approx |\Theta_{bilinear}| = d \times d_h + |\mathcal{V}|.
\end{align}
But since the parameters of $\Theta_{tied}$ are tied to the parameters of the input embedding, the effective number of parameters dedicated to the output layer is only $|\Theta_{tied}| =  |\mathcal{V}|$.

The capacities above depend on \textit{external} factors, that is $|\mathcal{V}|$, $d$ and $d_h$, which affect not only the output layer parameters but also those of other parts of the network. In practice, for $\Theta_{base}$ the capacity $d_h$ can be controlled with an additional linear projection on top of $h_t$  (e.g.~as in the OpenNMT implementation), but even in this case the parametrization would still be heavily dependent on $|\mathcal{V}|$.
Thus, the following inequality for the effective capacity of these models holds true for fixed $|V|$, $d$, $d_{h}$:
\begin{align}
\mathcal{C}_{tied} < \mathcal{C}_{bilinear} < \mathcal{C}_{base}.
\label{capacity_ineq}
\end{align}
This creates in practice difficulty in choosing the optimal capacity of the output layer which scales to large vocabularies and avoids underparametrization or overparametrization (left and right side of Eq.~\ref{capacity_ineq} respectively). Ideally, we would like to be able to choose the effective capacity of the output layer more flexibly moving freely in between $C_{bilinear}$ and $C_{base}$ in Eq.~\ref{capacity_ineq}.

\section{Structure-aware Output Layer for Neural Machine Translation}
The proposed \textit{structure-aware} output layer for neural machine translation, noted as NMT-\code{joint}, aims to learn the structure of the output space by learning a joint embedding between translation contexts and output classifiers, as well as, by learning what to share with input embeddings (Figure \ref{fig:proposed}). In this section, we describe the model in detail, showing how it can be trained efficiently for arbitrarily high number of effective parameters and how it is related to weight tying.

\subsection{Joint Input-Output Embedding}
  Let $g_{inp}(h_{t})$ and $g_{out}(e_{j})$ be two non-linear projections of $d_{j}$ dimensions of any translation context $h_{t}$ and any embedded output $e_{j}$, where $e_{j}$ is the j$_{th}$ row vector from the input embedding matrix E, which have the following form:
\begin{gather}
e_{j}' = g_{out}(e_{j}) = \sigma(U e_{j}^T + b_{u})\\
h_t' = g_{inp}(h_{t}) = \sigma( V h_{t} + b_{v}),
\end{gather}
\noindent where the matrix $U \in \rm I\!R^{d_j \times d}$ and bias $b_{u} \in \rm I\!R^{d_j}$ is the linear projection of the translation context and the matrix $V \in \rm I\!R^{ d_j \times  d_h}$ and bias $b_{v} \in \rm I\!R^{d_j}$ is the linear  projection of the outputs, and $\sigma$ is a non-linear activation function (here we use \code{Tanh}). Note that the projections could be high-rank or low-rank for $h_t'$ and $e'_{j}$ depending on their initial dimensions and the target joint space dimension.

With $E' \in \rm I\!R^{|\mathcal{V}| \times d_{j}} $ being the matrix resulting from projecting all the outputs $e_j$ to the joint space, i.e. $g_{out}(E)$, and a vector $b \in \rm I\!R^{|\mathcal{V}|}$ which captures the bias for each output, the conditional output probability distribution of Eq~\ref{baseline_eq} can be re-written as follows:
\begin{align}
  p&(y_t|y_1^{t-1},X) \label{jointeq} \\
  & \propto    \text{exp}\big( E' h_t' +  b\big)	 \nonumber\\
  & \propto  \text{exp}\big( g_{out}(E) g_{inp}(h_t) + b\big) \nonumber\\
  & \propto  \text{exp}\big( \sigma(U E^T + b_{u})\,
  \sigma( V h_{t} + b_{v}) + b\big).\nonumber
\end{align}
\subsubsection{What Kind of Structure is Captured?}
From the above formula we can derive the general form of the joint space which is similar to Eq.~\ref{nonlin} with the difference that it incorporates both  components for learning output and context structure:
\begin{equation}
\underbrace{\sigma(E \mathcal{W}_{o})}_{\text{Output structure~~~}} \!\!\!\!   \underbrace{\sigma( \mathcal{W}_{c} h_{t} )}_{\text{~~~Context structure}},
\label{oursgen}
\end{equation}
where $\mathcal{W}_{o} \in \rm I\!R^{d \times d_j}$ and $\mathcal{W}_{c} \in \rm I\!R^{d_j \times d_h}$ are the dedicated projections for learning output and context structure respectively (which correspond to $U$ and $V$ projections in Eq.~\ref{jointeq}).
We argue that both nonlinear components are essential and validate this hypothesis empirically in our evaluation by performing an ablation analysis (Section \ref{ablation}).

\subsubsection{How to Control the Effective Capacity?}

The capacity of the model in terms of effective number of parameters ($\Theta_{joint}$) is:
\begin{align}
 \mathcal{C}_{joint} \approx |\Theta_{joint}| = d \times d_j + d_j \times d_h + |\mathcal{V}|.
\end{align}
By increasing the joint space dimension $d_j$ above, we can now move freely between $\mathcal{C}_{bilinear}$ and $\mathcal{C}_{base}$ in Eq~.\ref{capacity_ineq} without depending anymore on the external factors ($d$, $d_h$, $|V|$) as follows:
\begin{align}
	\mathcal{C}_{tied} < \mathcal{C}_{bilinear} \leq \mathcal{C}_{joint}  \leq \mathcal{C}_{base}.
\end{align}
However, for very large number of $d_j$ the computational complexity increases prohibitively because the projection requires a large matrix multiplication between $U$ and $E$ which depends on $|\mathcal{V}|$. In such cases, we resort to sampling-based training, as explained in the next subsection.
\begin{table*}[htp]
	\center
	\small
	{\def\arraystretch{1.5}\tabcolsep=3pt
		\begin{tabular}{p{1em}l ll ll ll ll}  \hline
      &  &\multicolumn{2}{c}{\textbf{En $\rightarrow$ Fi}} & \multicolumn{2}{c}{\textbf{Fi $\rightarrow$ En}}  &\multicolumn{2}{c}{\textbf{En $\rightarrow$ De}} & \multicolumn{2}{c}{\textbf{De $\rightarrow$ En}} \\
		& \textbf{Model} &\textbf{$|\Theta|$} & {BLEU} ($\Delta$)  & \textbf{$|\Theta|$}  & {BLEU} ($\Delta$)  &\textbf{$|\Theta|$}  & {BLEU} ($\Delta$)   & \textbf{$|\Theta|$} & {BLEU} ($\Delta$)  \\ \hline\hline
     \parbox[t]{1mm}{\multirow{3}{*}{\rotatebox[origin=c]{90}{  {32K}  }}} & NMT  & 60.0M &  12.68 (--)  & 59.8M & 9.42  (--) & 61.3M &  18.46  (--) & 65.0M & 15.85  (--)  \\
	 &	NMT-\code{tied}  & 43.3M  & 12.58 (\textcolor{black}{$-0.10$})  & 43.3M  & 9.59 ($+0.17$) &   44.9M &  18.48 ($+0.0$)   & 46.7M & 16.51  ($+0.66$)$\dagger$  \\
	 &	NMT-\code{joint}   & 47.5M & \textbf{13.03} ($+0.35$)$\ddagger$    & 47.5M  &  \textbf{10.19} ($+0.77$)$\ddagger$  & 47.0M &  \textbf{19.79}   ($+1.3$)$\ddagger$   & 48.8M & \textbf{18.11}  ($+2.26$)$\ddagger$  \\
    \hline
  \parbox[t]{1mm}{\multirow{3}{*}{\rotatebox[origin=c]{90}{  {64K}  }}} &  NMT   & 108.0M &  13.32  (--)  & 106.7M &  \textbf{12.29}  (--) &  113.9M  &  20.70  (--) &  114.0M & 20.01  (--) \\
  &  NMT-\code{tied}   & 75.0M & 13.59 ($+0.27$)  & 75.0M & 11.74 (\textcolor{black}{$-0.55$})$\ddagger$  &  79.4M &  20.85 (\textcolor{black}{$+0.15$})  &  79.4M &  19.19 (\textcolor{black}{$-0.82$})$\dagger$ \\
  &  NMT-\code{joint} & 75.5M & \textbf{13.84}  ($+0.52$)$\ddagger$   & 75.5M & 12.08 ($-0.21$) & 79.9M  & \textbf{21.62}  ($+0.92$)$\ddagger$  & 79.9M & \textbf{20.61} ($+0.60$)$\dagger$  \\ \hline 
	\parbox[t]{1mm}{\multirow{3}{*}{\rotatebox[origin=c]{90}{  {128K}	($\sim$)  }}} &  NMT  & 201.1M &  13.52 (--)  & 163.1M & 11.64 (--) & 211.3M  &  22.48 (--)  & 178.3M  & 19.12 (--) \\
  &  NMT-\code{tied}  & 135.6M & {13.90} ($+0.38$)$^*$   & 103.2M  & 11.97 ($+0.33$)$^*$ &  144.2M &  21.43 ($-0.0$) &  111.6M & 19.43 ($+0.30$) \\
  &  NMT-\code{joint} & 137.7M & \textbf{13.93} ($+0.41$)$\dagger$   & 103.7M  & \textbf{12.07} ($+0.43$)$\dagger$  & 146.3M  & \textbf{22.73} ($+0.25$)$\dagger$  & 115.8M  & \textbf{20.60} ($+1.48$)$\ddagger$ \\ \hline
	\end{tabular}}
\caption{Model performance and number of parameters ($|\Theta|$) with varying BPE operations (32K, 64K, 128K) on the English-Finish and English-German language pairs. The significance of the difference against the NMT baseline with $p$-values $<$$.05$, $<$$.01$ and $<$$.001$ are marked with $^*$, $\dagger$ and $\ddagger$ respectively.}
	\label{tab:scores}
\end{table*}

\subsection{Sampling-based Training}
To scale up to large output sets we adopt the negative sampling approach from \cite{mikolov2013distributed}. The goal is to utilize only a sub-set $\mathcal{V}'$ of the vocabulary instead of the whole vocabulary $\mathcal{V}$ for computing the softmax. The sub-set $\mathcal{V}'$ includes all positive classes whereas the negative classes are randomly sampled. During back propagation only the weights corresponding to the sub-set $\mathcal{V}'$ are updated. This can be trivially extended to mini-batch stochastic optimization methods by including all positive classes from the examples in the batch and sampling negative examples randomly from the rest of the vocabulary.

Given that the joint space models generalize well on seen or unseen outputs \cite{yazdani2015model,Nam16}, we hypothesize that the proposed joint space will be more sample efficient than the baseline NMT with or without \textit{weight tying}, which we empirically validate with a sampling-based experiment in Section \ref{emb_size_effect} (Table \ref{tab:scores}, last three rows with $|\mathcal{V}| \approx 128K$).

\subsection{Relation to \textit{Weight Tying}}
The proposed joint input-output space can be seen as a generalization of \textit{\textit{weight tying}} ($W=E^T$, Eq.~\ref{baseline_eq}), because its degenerate form is equivalent to \textit{weight tying}. In particular, this can be simply derived if we set the non-linear projection functions in the second line of Eq.~\ref{jointeq} to be the identity function, $g_{inp}(\cdot) = g_{out}(\cdot) = I$, as follows:
\begin{align}
p(y_t|y_1^{t-1},X) 
   & \propto  \text{exp}\big( (I E) \  (I h_{t}) + b\big) \nonumber \\
    & \propto  \text{exp}\big(  E  h_{t} + b\big)\ \square.
\label{gen}
\end{align}
Overall, this new parametrization of the output layer generalizes over previous ones and addresses their aforementioned challenges in Section \ref{challenges}.

\section{Evaluation}
\vspace{-2mm}
We compare the NMT-\code{joint} model to two strong NMT baselines trained with and without \textit{weight tying} over four large parallel corpora which include morphologically rich languages as targets (Finnish and German), but also morphologically less rich languages as targets (English) from WMT 2017 \cite{bojar-EtAl:2017:WMT1}\footnote{\url{http://www.statmt.org/wmt17/}}. We  examine the behavior of the proposed model under challenging conditions, namely varying vocabulary sizes, architecture depth, and output frequency.

\subsection{Datasets and Metrics}
The English-Finnish corpus contains 2.5M sentence pairs for training, 1.3K for development (Newstest2015), and 3K for testing (Newstest2016), and the English-German corpus 5.8M for training,  3K for development (Newstest2014), and 3K  for testing (Newstest2015). We pre-process the texts using the BPE algorithm  \cite{sennrich-haddow-birch:2016:P16-12} with 32K, 64K and 128K operations. Following the standard evaluation practices in the field \cite{bojar-EtAl:2017:WMT1}, the translation quality is measured using BLEU score  \citep{papineni-EtAl:2002:ACL} (\emph{multi-blue}) on \emph{tokenized} text and the significance is measured with  the  paired  bootstrap re-sampling method proposed by  \cite{koehn-EtAl:2007:PosterDemo}.\footnote{\texttt{multi-bleu.perl} and \texttt{bootstrap-hypothe- sis-difference-significance.pl} scripts.}
The quality on infrequent words is measured with METEOR \cite{denkowski:lavie:meteor-wmt:2014} which has  originally been proposed to measure performance on function words. To adapt it for our purposes on English-German pairs ($|\mathcal{V}| \approx 32K$), we set as \emph{function words} different sets of words grouped according to three frequency bins, each of them containing  $\frac{|\mathcal{V}|}{3}$ words of \textit{high}, \textit{medium} and  \textit{low} frequency respectively and set its parameters to $\{0.85, 0.2, 0.6, 0.\}$  and  $\{0.95, 1.0, 0.55, 0.\}$  when evaluating on English and German respectively.
\vspace{0mm}

\subsection{Model Configurations}
The baseline is an encoder-decoder with 2 stacked LSTM layers on each side from OpenNMT \cite{klein-EtAl:2017:ACL-2017-System-Demonstrations}, but we also experiment with varying depth in the range $\{1, 2, 4, 8\}$ for German-English. The hyperparameters are set according to validation accuracy as follows: maximum sentence length of  50, 512-dimensional word embeddings and LSTM hidden states, dropout with a probability of 0.3 after each layer, and Adam \cite{kingma2014adam} optimizer with initial learning rate of 0.001. The size of the joint space is also selected on validation data in the range $\{512, 2048, 4096\}$. For efficiency, all models on corpora with $\mathcal{V}\approx 128K$ ($\sim$) and all \textit{structure-aware}  models with $d_j \geq 2048$  on corpora with $\mathcal{V} \leq 64K$ are trained with 25\% negative sampling.\footnote{Training the models with a full 128K vocabulary without sampling runs out of memory on our machines.}

\subsection{Translation Performance}
Table \ref{tab:scores} displays the results on four translation sets from English-German and English-Finish language pairs when varying the number of BPE operations.  The NMT-\code{tied} model outperforms the NMT baseline in many cases, but the differences are not consistent and it even scores significantly lower than NMT baseline in two cases, namely on Fi $\rightarrow$ En and De $\rightarrow$ En with $\mathcal{V}\approx 64K$. This validates our claim that the parametrization of the output space of the original NMT is not fully redundant, otherwise the NMT-\code{tied} would be able to match its BLEU in all cases. In contrast, the NMT-\code{joint} model outperforms consistently both baselines with a difference up to $+$2.2 and $+$1.6 BLEU points respectively,\footnote{Except in the case of Fi $\rightarrow$ En with $|\mathcal{V}| \approx 64K$, where the NMT baseline performed the best.} showing that the NMT-\code{tied} model has a more effective parametrization and retains the advantages of both baselines, namely sharing weights with the input embeddings, and dedicating enough parameters for generation.

\begin{table}
 \centering
 	{\def\arraystretch{1.25}\tabcolsep=1.3pt
\begin{tabular}{l lllc }\hline
 & \textbf{Model} & \textbf{Layer form} & \textbf{BLEU} & $|\Theta|$  \\\hline
	& NMT & $W^Th_{t}$ &   15.85 &65.0M  \\
	& NMT-\code{tied} & $E h_{t}$ &  16.51 & 46.7M\\\hdashline
 \parbox[t]{3mm}{\multirow{5}{*}{\rotatebox[origin=c]{90}{  {NMT-\code{joint}}  }}} &  Eq.~\ref{eq:bilinear}	& $E \mathcal{W}  h_{t} $ &  16.23 & 47.0M \\
 & Eq.~\ref{nonlin} a 	    & $\sigma(E \mathcal{W})  h_{t} $ & 16.01  & 47.0M\\
 & Eq.~\ref{nonlin} b 	& $E \sigma(\mathcal{W}  h_{t}) $ & 17.52 & 47.0M\\
 & Eq.~\ref{oursgen} (512) & $\sigma(E \mathcal{W}_{o}) \sigma( \mathcal{W}_{c} h_{t} )$ &  17.54  & 47.2M\\
 & Eq.~\ref{oursgen} (2048) & $\sigma(E \mathcal{W}_{o}) \sigma( \mathcal{W}_{c} h_{t} )$ &  \textbf{18.11}  & 48.8M\\ \hline
\end{tabular}}
\caption{BLEU scores on De $\rightarrow$ En ($|\mathcal{V}| \approx 32K$) for the ablation analysis of NMT-\code{joint}.}
\label{ablationres}
\vspace{-5mm}
\end{table}

Overall, the highest scores correlate with a high number of BPE operations, namely 128K, 64K, 128K and 64k respectively. This suggests that the larger the vocabulary the better the performance, especially for the morphologically rich target languages, namely En $\rightarrow$ Fi and En $\rightarrow$ De.
Lastly, the NMT baseline seems to be the least robust to sampling since its BLEU decreases in two cases. The other two models are more robust to sampling, however the difference of NMT-\code{tied} with the NMT is less significant than that of NMT-\code{joint}.

\subsection{Ablation Analysis}
\label{ablation}
To demonstrate whether all the components of the proposed joint input-output model are useful and to which extend they contribute to the performance, we performed an ablation analysis; the results are displayed in Table \ref{ablationres}. Overall, all the variants of the NMT-\code{joint} outperform the baseline with varying degrees of significance. The NMT-\code{joint} with a bilinear form (Eq.~~\ref{eq:bilinear}) as in \cite{yazdani2015model,Nam16} is slightly behind the NMT-\code{tied} and outperforms the NMT baseline; this supports our theoretical analysis in Section \ref{weighttying} which demonstrated that \textit{weight tying} is learning an implicit linear structure similar to bilinear joint input-output models.

\begin{figure}\vspace{-4mm}
  \centering
  \hspace{-3mm}\includegraphics[width=0.49\textwidth]{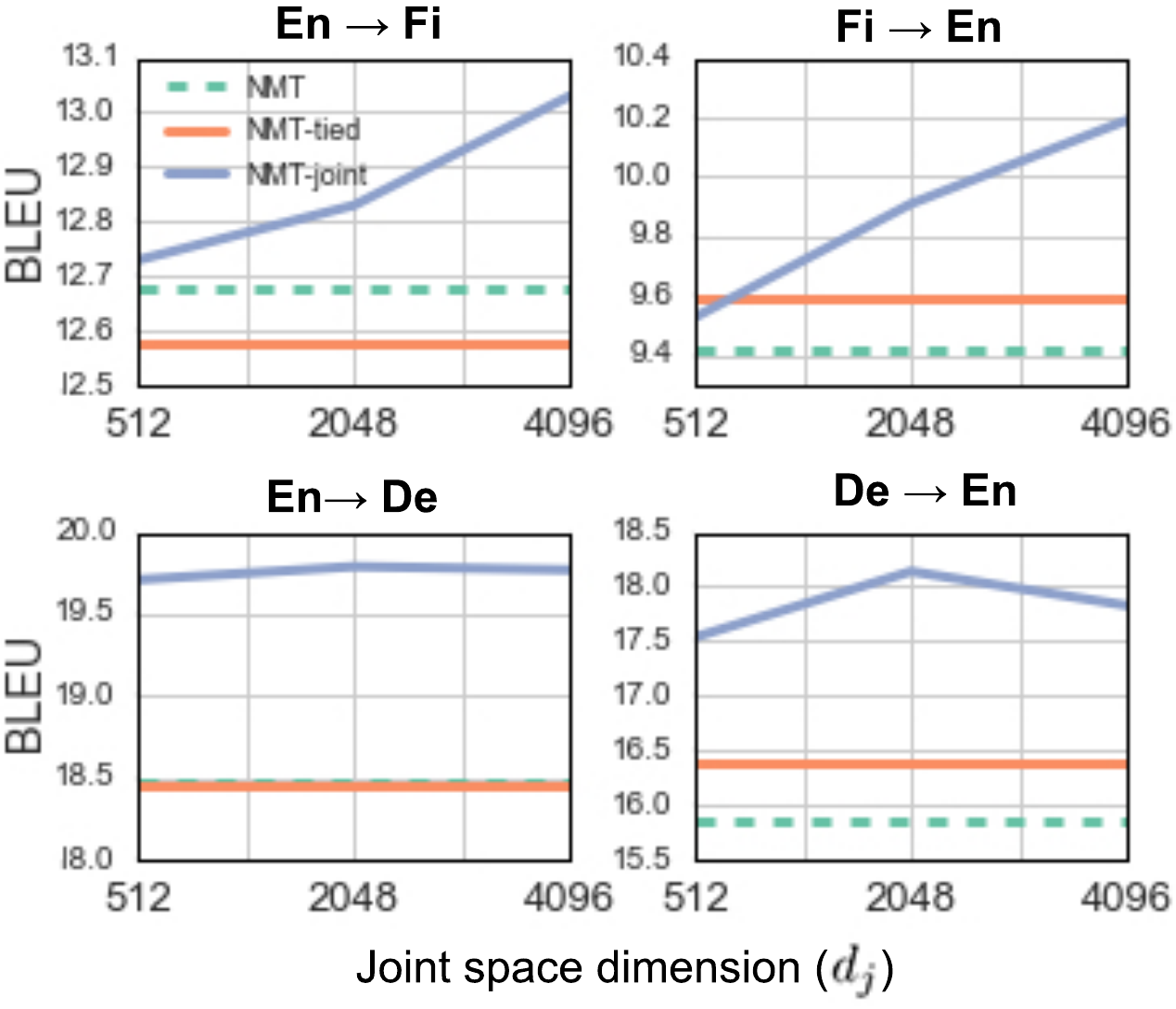}
	\vspace{-3mm}
  \caption{BLEU scores for the NMT-\code{joint} model when varying its dimension ($d_j$) with $|\mathcal{V}| \approx 32K$. }
	\label{fig:emb_size}
	\vspace{-3mm}
\end{figure}

The NMT-\code{joint} model without learning explicit translation context structure (Eq.~\ref{nonlin} a) performs similar to the bilinear model and the NMT-\code{tied} model, while the NMT-\code{joint} model without learning explicit output structure (Eq.~\ref{nonlin} b) outperforms all the previous ones.
When keeping same capacity (with $d_{j}{=}512$), our full model, which learns both output and translation context structure, performs similarly to the latter model and
outperforms all the other baselines, including joint input-output models with a bilinear form \cite{yazdani2015model,Nam16}.  But when the capacity is allowed to increase (with $d_{j}{=}2048$), it outperforms all the other models.  Since both nonlinearities are necessary to allow us to control the effective capacity of the joint space, these results show that both types of structure induction are important for reaching the top performance with NMT-\code{joint}.
\begin{figure*}[htbp]
  \centering
    \begin{subfigure}[b]{0.47\textwidth}
      \centering
        \includegraphics[width=0.84\textwidth]{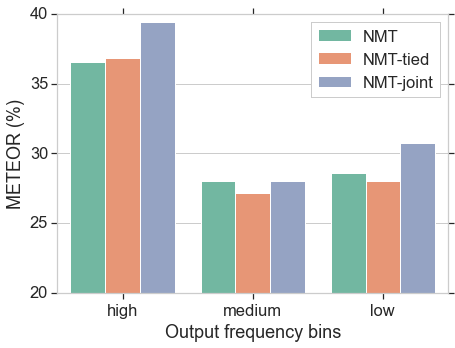}
        \vspace{-1mm}\caption{Results on En $\rightarrow$ De ($|\mathcal{V}| \approx 32K$). }
        \label{fig:tied}
    \end{subfigure}
		\hspace{3mm}
    \begin{subfigure}[b]{0.47\textwidth}
      \centering
        \includegraphics[width=0.84\textwidth]{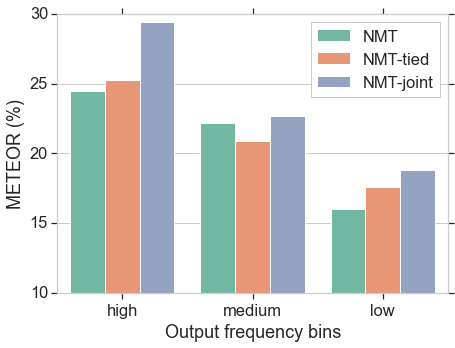}
        \vspace{-1mm}\caption[1]{Results on De $\rightarrow$ En ($|\mathcal{V}| \approx 32K$). }
     \end{subfigure}
\vspace{-2mm}
    \caption[1]{METEOR scores (\%) on both directions of German-English language pair for all the models when focusing the evaluation on different frequency outputs grouped into three bins (high, medium, low).}
		\label{fig:outfreq}
\vspace{-2mm}
\end{figure*}

\subsection{Effect of Embedding Size}
\label{emb_size_effect}

\textbf{Performance} Figure \ref{fig:emb_size} displays the BLEU scores of the proposed model when varying the size of the joint embedding, namely $d_{j} \in \{512, 2048, 4096\}$, against the two baselines. For English-Finish pairs, the increase in embedding size leads to a consistent increase in BLEU in favor of the NMT-\code{joint} model. For the English-German pairs, the difference with the baselines is much more evident and the optimal size is observed around 2048 for De $\rightarrow$ En and around 512 on En $\rightarrow$ De. The results validate our hypothesis that there is parameter redundancy in the typical output layer.  However the ideal parametrization is data dependent and is achievable systematically only with the \code{joint} output layer which is capacity-wise in between the typical output layer and the \code{tied} output layer.

\begin{table}
	\centering
\begin{tabular}{ ll llll }\hline
&	  & \multicolumn{3}{c}{{   \textbf{Sampling }   }}\\
\textbf{Model} & $d_j$ & 50\%   & 25\%  & 5\%   \\ \hline\hline
  NMT  & - & 4.3K  & 5.7K & 7.1K   \\
	NMT-\code{tied} & - & 5.2K &  6.0K & 7.8K \\
  NMT-\code{joint} &  512 & 4.9K  & 5.9K & 7.2K\\
  NMT-\code{joint} & 2048 & 2.8K  & 4.2K & 7.0K\\
	NMT-\code{joint} & 4096 & 1.7K  & 2.9K & 6.0K\\ \hline
\end{tabular}
\caption{Target tokens processed per second during training with negative sampling on En $\rightarrow$ De pair with a large BPE vocabulary  $|\mathcal{V}| \approx 128K$.}
\label{speed}
\vspace{-4mm}
\end{table}

\textbf{Training speed} Table \ref{speed} displays the target tokens processed per second by the models on En $\rightarrow$ DE with $|\mathcal{V}|\approx 128K$ using different levels of negative sampling, namely 50\%, 25\%, and 5\%. In terms of training speed, the 512-dimensional NMT-\code{joint} model is as fast as the baselines, as we can observe in all cases. For higher dimensions of the joint space, namely 2048 and 4096 there is a notable decrease in speed which is remidiated by reducing the percentage of the negative samples.

\begin{table*}
 \centering
\begin{tabular}{ c ccccccccc }\hline
 \textbf{Model} & $d_j$ & \textbf{1-layer} & $|\Theta|$ & \textbf{2-layer}  & $|\Theta|$ & \textbf{3-layer}  & $|\Theta|$ & \textbf{4-layer}  & $|\Theta|$ \\\hline
	NMT &   - & 16.49 & 60.8M & 15.85 & 65.0M & 17.71 & 69.2M & 17.74 & 73.4M \\
	NMT-\code{tied} &  - &15.93 & 42.5M & 16.51 & 46.7M  & 17.72 & 50.9M &  17.60 & 55.1M\\
	NMT-\code{joint}  &  512 &\textbf{16.93} & 43.0M & \textbf{17.54} & 47.2M  & \textbf{17.83} & 51.4M& \textbf{18.13} & 55.6M\\ \hline
\end{tabular}
\caption{BLEU scores on De $\rightarrow$ En ($|\mathcal{V}| \approx 32K$) for the NMT-\code{joint} with $d_j=512$ against baselines when varying the depth of both the encoder and the decoder of the NMT model.}
\label{depth}
\end{table*}
\subsection{Effect of Output Frequency and Architecture Depth}
Figure \ref{fig:outfreq} displays the performance in terms of METEOR on both directions of German-English language pair when evaluating on outputs of different frequency levels (high, medium, low) for all the competing models. The results on  De $\rightarrow$ EN show that the improvements brought by the NMT-\code{joint} model against baselines are present consistently for all frequency levels including the low-frequency ones. Nevertheless, the improvement is most prominent for high-frequency outputs, which is reasonable given that no sentence filtering was performed and hence frequent words have higher impact in the absolute value of METEOR. Similarly, for  En $\rightarrow$ De we can observe that NMT-\code{joint} outperforms the others on high-frequency and low-frequency labels while it reaches parity with them on the medium-frequency ones.

We also evaluated our model in another challenging condition in which we examine the effect of the NMT architecture depth in the performance of the proposed model. The results are displayed in Table \ref{depth}. The results show that the NMT-\code{joint} outperforms the other two models consistently when varying the architecture depth of the encoder-decoder architecture. The NMT-\code{joint} overall is much more robust than NMT-\code{tied} and it outperforms it consistently in all settings. Compared to the NMT which is overparametrized the improvement even though consistent it is smaller for layer depth 3 and 4. This happens because NMT has a much higher number of  parameters than NMT-\code{joint} with $d_j{=}512$.

Increasing the number of  dimensions $d_j$ of the joint space should lead to further improvements, as shown in Fig.~2. In fact, our NMT-\code{joint} with $d_j=2048$ reaches 18.11 score with a 2-layer deep model, hence it outperforms all other NMT and NMT-\code{tied} models even with a deeper architecture (3-layer and 4-layer) regardless of the fact that it utilizes fewer parameters than them (48.8M vs 69.2-73.4M and 50.9-55.1M respectively).

\section{Related Work}

Several studies focus on learning joint input-output representations grounded to word semantics for zero-shot image classification \citep{Weston2011,Socher13,ZhangGS16}, but there are fewer such studies for NLP tasks. \citep{yazdani2015model} proposed a zero-shot spoken language understanding model based on a bilinear joint space trained with hinge loss, and  \citep{Nam16}, proposed a similar joint space trained with a WARP loss for zero-shot biomedical semantic indexing. In addition, there exist studies which aim to learn output representations directly from data such as \citep{Srikumar14,YehWKW17,augenstein18}; their lack of semantic grounding to the input embeddings and the vocabulary-dependent parametrization, however, makes them data hungry and less scalable on large label sets.
All these models, exhibit similar  theoretical  limitations as the softmax linear unit with \textit{weight tying} which were described in Sections 2.2.

To our knowledge, there is no existing study which has considered the use of such joint input-output labels for neural machine translation. Compared to previous joint input-label models our model is more flexible and not restricted to linear mappings, which have limited expressivity, but uses non-linear mappings modeled similar to energy-based learning networks \citep{belanger16}. Perhaps, the most similar embedding model to ours is the one by \citep{pappas_18a}, except for the linear scaling unit which is specific to sigmoidal linear units designed for multi-label classification problems and not for structured prediction, as here.

\section{Conclusion and Perspectives}
We proposed a re-parametrization of the output layer for the decoder of NMT models which is more general and robust than a softmax linear unit with or without \textit{\textit{weight tying}}  with the input word embeddings. Our evaluation shows that the \textit{structure-aware} output layer outperforms \textit{\textit{weight tying}} in all cases and  maintains a significant difference with the typical output layer without compromising much the training speed. Furthermore, it can successfully benefit from training corpora with large BPE vocabularies using negative sampling. The ablation analysis demonstrated that both types of structure captured by our model are essential and complementary, as well as, that their combination outperforms all previous output layers including those of bilinear input-output embedding models. Our further investigation revealed the robustness of the model to sampling-based training, translating infrequent outputs and to varying architecture depth.

As future work, the \textit{structure-aware} output layer could be further improved along the following directions. The computational complexity of the model becomes prohibitive for a large joint projection because it requires a large matrix multiplication which depends on  $|\mathcal{V}|$; hence, we have to resort to sampling based training relatively quickly when gradually increasing $d_j$ (e.g.~for $d_j >= 2048$). A more scalable way of increasing the output layer capacity could address this issue, for instance, by considering multiple consecutive additive transformations with small $d_j$. Another useful direction would be to use more advanced output encoders and additional external knowledge (contextualized or generically defined) for both words and sub-words. Finally, to encourage progress in joint input-output embedding learning for NMT, our code is available on Github: \url{http://github.com/idiap/joint-embedding-nmt}.

\section*{Acknowledgments}

We are grateful for the support from the European Union through its Horizon 2020 program in the SUMMA project n.\ 688139, see \url{http://www.summa-project.eu} and for the valuable feedback from the anonymous reviewers.

\bibliography{references}
\bibliographystyle{acl_natbib_nourl}

\appendix

\end{document}